%% file: main.tex
\documentclass[10pt,twocolumn,letterpaper]{article}
\usepackage[accsupp]{axessibility}  
\usepackage[pagenumbers]{wacv}  

\usepackage{multirow}
\usepackage{xcolor}
\definecolor{wacvblue}{rgb}{0.21,0.49,0.74}
\usepackage[pagebackref,breaklinks,colorlinks,allcolors=wacvblue]{hyperref}

\newif\iftestcerpending \testcerpendingfalse
\newcommand{\tc}[1]{\iftestcerpending{\color{red}#1}\else#1\fi}

\newcommand{\cer}{CER}
\newcommand{\zi}{0-iter}


\def\wacvPaperID{*****}
\def\confName{WACV}
\def\confYear{2027}

\title{Dataset Distillation for Handwritten Text Recognition:\\An Empirical Study}

\author{Félix Fuentes-Hurtado\\
University of Alicante, Spain\\
{\tt\small felix.fuentes@ua.es}
\and
Eric Ayllón\\
University of Alicante, Spain\\
{\tt\small eric.ayllon@ua.es}
\and
Juan C. Alfaro\\
University of Castilla-La Mancha, Spain\\
{\tt\small juancarlos.alfaro@uclm.es}
}

\begin{document}
\maketitle
\input{sections/abstract}
\input{sections/intro}
\input{sections/related}
\input{sections/method}
\input{sections/setup}
\input{sections/results}
\input{sections/discussion}
\input{sections/conclusion}
\input{sections/ack}
{
    \small
    \bibliographystyle{ieeenat_fullname}
    \bibliography{main}
}

\end{document}

%% file: sections/abstract.tex
\begin{abstract}
Dataset distillation (DD) replaces large training datasets with smaller synthetic ones to reduce computational costs. While traditionally used for classification, we present, to our knowledge, the first DD study for line-level handwritten text recognition (HTR). Unlike classification tasks, which use discrete labels, HTR employs unaligned sequences from a virtually infinite output space as labels. To adapt classification-based DD methods to this scenario, we define text pseudo-classes by clustering transcriptions based on character n-grams. We adapt gradient matching (GM), distribution matching (DM) and trajectory matching (MTT) to a Convolutional Recurrent Neural Network (CRNN) trained with Connectionist Temporal Classification (CTC). In this process, synthetic line images are generated by optimising only the pixels of real starting lines (either randomly chosen or selected as text medoids), while keeping their original transcriptions fixed. Across four datasets (IAM, George Washington, READ\,2016, LAM) and three budgets (1--10\,\% of the training lines), no distilled set outperforms its real initialisation lines. Random lines prove better than medoids, and further optimisation only degrades recognition. Methodologically, DM lowers its objective without benefit, while GM and MTT barely move theirs. We turn these findings into guidelines for evaluating data compression in HTR. Code is available at \url{https://github.com/ffuhu/htr-dataset-distillation}.
\end{abstract}

%% file: sections/intro.tex
\section{Introduction}
\label{sec:intro}
Handwritten documents make up a large part of historical records, and most of them exist only in physical form, where
they are exposed to loss as they deteriorate. Scanning them is not enough to make them searchable, and manual
transcription is slow and expensive. Handwritten Text Recognition (HTR) automates transcription and lets digitisation
scale with minimal human intervention.

HTR models are routinely retrained to adapt to new collections, hands, or periods~\cite{muehlberger2019transkribus,kohut2023finetuning}. However, repeatedly processing full training sets becomes computationally prohibitive in routine workflows: finding optimal architectures and hyperparameters through multiple training cycles, revisiting past collections to prevent catastrophic forgetting when adapting to new domains, etc. Condensing large datasets into compact, highly informative synthetic sets would alleviate this burden, enabling rapid model exploration and efficient knowledge retention at a fraction of the cost.
Dataset
distillation (DD) pursues this goal by \emph{optimising} a small set of synthetic samples so that a model trained on
them performs close to one trained on the full collection. Three representative methods are Gradient Matching
(GM)~\cite{zhao2021dc}, Distribution Matching (DM)~\cite{zhao2023dm} and Matching Training Trajectories (MTT)~\cite{cazenavette2022mtt}.
On image classification these methods report large gains over random or
coreset selection at small budgets~\cite{zhao2021dc,cazenavette2022mtt,cui2022dcbench}. They rely, however, on each sample having a single class label: synthetic images are initialised, matched and labelled by class.

HTR lacks this property as a text line has no class: its label is a variable-length character sequence. To our knowledge, there is no prior work on DD for HTR. To bridge this gap in the literature, we adapt GM, DM and MTT to HTR. To do so, we generate synthetic line images by optimising only the pixels of real starting lines, keeping their original transcriptions and widths strictly fixed. Since this optimisation requires initial real images, we evaluate two starting setups. For the first approach all three methods start from a subset of \emph{random lines}. For the second approach, we group similar transcriptions to define \emph{text pseudo-classes} and start each synthetic line image from the \emph{medoid} of its respective pseudo-class.
We show that no method improves upon its initialisation or beats random real lines of the same size.

The main contribution of this work is an adaptation of GM, DM, and MTT to CTC-based line-level HTR, including the proposed pseudo-class method for conditional matching. We also perform an ablation study to discern whether final accuracy is due to initialisation or optimisation of DD methods. With the evidence we gather from the experiments (fragile second-order CTC gradients, step size \vs pixel range, random-subset variance, GPU effects), we elaborate a comprehensive set of guidelines for the evaluation of DD for HTR to facilitate future works on this challenging avenue.

%% file: sections/related.tex
\section{Related work}
\label{sec:related}

\paragraph{Dataset distillation.}
DD was introduced as bi-level optimisation of synthetic images through unrolled training~\cite{wang2018dataset}.
Gradient matching (GM, introduced as DC) matches the gradients of real and synthetic batches class by class~\cite{zhao2021dc}, Differentiable Siamese Augmentation (DSA) adds differentiable
augmentation shared by both the real and synthetic branches~\cite{zhao2021dsa}, DM matches mean embeddings under randomly initialised
networks~\cite{zhao2023dm}, and MTT matches the parameters reached by a student trained on synthetic data to those of
experts trained on real data~\cite{cazenavette2022mtt}.
Later work learns or relabels the targets~\cite{bohdal2020flexible,sucholutsky2021soft,cui2023tesla,yin2023sre2l},
re-parameterises the synthetic data~\cite{kim2022idc} or optimises in a generator's latent
space~\cite{cazenavette2023glad}; see~\cite{yu2024review,sachdeva2023survey,geng2023survey}. Beyond images, DD has been applied to text
classification~\cite{sucholutsky2021soft}, graphs~\cite{jin2022gcond}, spoken keyword classification~\cite{kim2022idc}
and speech emotion recognition~\cite{ritter2024speech}. In all of these each sample carries a class (or a soft class
distribution); we found no work with unaligned sequence labels or a CTC objective.

\paragraph{Critical evaluations of DD.}
DC-BENCH~\cite{cui2022dcbench} is closest in spirit to our study: under a common protocol it showed that augmentation at
evaluation time boosts all methods, that the advantage over random selection vanishes as images per class grow, and
that initialisation matters. The advantage over random subsets also depends on soft labels~\cite{dey2026hardtruths},
and accuracy alone is a questionable metric~\cite{li2025ddranking}.

\paragraph{Coreset selection} chooses real samples by herding~\cite{welling2009herding}, $k$-center
covering~\cite{sener2018coreset} or forgetting statistics~\cite{toneva2019forgetting}, among others~\cite{guo2022deepcore};
our text medoids are a cheap selector on transcriptions only.

\paragraph{Data-efficient HTR.}
HTR has become cheaper to train through lighter annotation requirements and more efficient models, but systems are still trained or fine-tuned on whole training splits. Some work on data efficiency for HTR has been done to reduce the number of \emph{labelled} lines required. Aradillas et al.~\cite{aradillas2021boosting} studied few-shot learning. However, other efforts such as synthetic handwriting~\cite{kang2020ganwriting,bhunia2021hwt} or augmentation~\cite{wigington2017augmentation} fail to reduce the actual number of samples fed during training or fine-tuning. Knowledge distillation for text recognition~\cite{zhang2024dctc} and speech recognition~\cite{takashima2018kd} compresses \emph{models}. In contrast to this model-centric approach, we apply DD to study whether the \emph{training data} itself can be compressed.

%% file: sections/method.tex
\section{Method}
\label{sec:method}

\subsection{Problem setting}
\label{sec:setting}
Let $\mathcal{D}=\{(x_i,y_i)\}_{i=1}^{N}$ be a pool of text-line images $x_i\in\mathbb{R}^{H\times W_i}$ 
(with fixed height $H$ and variable width $W_i$) with transcriptions $y_i\in\Sigma^{*}$
over a character alphabet $\Sigma$. For a given budget $\rho$, we build a set
$\mathcal{S}=\{(s_j,\tilde y_j)\}_{j=1}^{K}$ with $K=\rho N$ (\cref{tab:datasets}) and train an HTR model from
scratch on $\mathcal{S}$ alone. A set is either $K$ real lines (a \emph{selection}) or the result of a DD method that
starts from $K$ real lines and optimises their pixels. Each synthetic line inherits the transcription and the width of
the real line it starts from, both of which stay fixed: $\tilde y_j=y_{\pi(j)}$, where $\pi(j)$ is the index of the
starting line, and only the pixels $s_j$ are learned.

\subsection{HTR}
\label{sec:crnn}
To model the inherent mapping from images to their transcriptions $f: \mathcal{X} \rightarrow \Sigma^*$, we use a Convolutional Recurrent Neural Network (CRNN)~\cite{shi2017crnn,puigcerver2017mdlstm} trained with the Connectionist Temporal Classification (CTC) loss~\cite{graves2006ctc}. The CRNN is composed of a sequence of convolutional blocks and max-pooling layers that extract visual features and reduce the spatial dimensions. The resulting $T$ temporal frames are fed into a bidirectional LSTM and then a linear layer to project the resulting sequence to $|\Sigma|+2$ to account for all characters, including the blank and padding symbols. Early in training, CTC is attracted to a degenerate solution in which almost every frame is labelled blank, and the output bias reinforces it~\cite{zeyer2021peaky}. Following the prior-based bias initialisation of~\cite{lin2017focal}, we initialise the blank's output bias to $-5$ (chosen empirically). Exact architecture and hyperparameters are detailed in \cref{sec:protocol}.

\subsection{Starting lines and conditioning}
\label{sec:init}
Since we generate synthetic data by modifying real images, every DD method requires an initial set of $K$ starting lines. We evaluate  two initialization strategies. To isolate the effect of pixel optimisation, the original unaltered images of these starting lines also serve as selection baselines. In addition, we also try both \textit{unconditional} and \textit{conditional} matching, using pseudo-classes from the text medoids in the DD methods that allow for it (GM, DM).

\emph{Random lines.} $K$ lines drawn uniformly from the pool, using a different subset for each random seed. All the three methods can be initialised this way. When trained directly without optimisation, this set constitutes the \emph{Random} baseline.

\emph{Text medoids.} GM and DM match real and synthetic data class by class in image
classification~\cite{zhao2021dc,zhao2023dm}. Since HTR lines lack discrete classes, we define $K$ \emph{text pseudo-classes} by
clustering transcriptions. Each $y_i$ is represented by TF-IDF weights of its character 2- and 3-grams, and $k$-means
with $k=K$ partitions the pool into clusters $\mathcal{C}_1,\dots,\mathcal{C}_K$. The medoid of
$\mathcal{C}_j$ is the member closest to the cluster centroid in TF-IDF space. The synthetic line $j$ is initialised with the
medoid of $\mathcal{C}_j$ and is matched against real lines from $\mathcal{C}_j$ (\emph{conditional} matching). Trained
without optimisation, these $K$ medoids are the \emph{Text medoids} baseline, effectively acting as a cheap selector based solely on transcriptions. Note that MTT matches training trajectories rather than classes, so it is initialised from random lines only.

Since random lines carry no pseudo-class, GM and DM initialised this way match against real lines sampled from the entire pool
(\emph{unconditional} matching). We denote the methods started from random lines as DM and GM, and those started from medoids as
DM$_\mathrm{med}$ and GM$_\mathrm{med}$. Finally, to isolate the effect of conditional matching from the choice of starting lines, we also evaluate DM$_\mathrm{med}$ and GM$_\mathrm{med}$ unconditionally (\cref{sec:init-matters}).

\emph{Untrained controls.} A DD method initialises a synthetic set of real lines and subsequently optimises their pixels. Crucially, distilled sets are stored and loaded directly as tensors, whereas real lines are traditionally loaded as images. Simply comparing a distilled set with a standard selection of real lines would therefore conflate the effect of pixel optimisation with the effect of the loading pipeline. To disentangle these two factors, for each set of starting lines, we establish an untrained control by passing the exact same lines through the tensor pipeline with zero optimisation steps (\emph{\zi}). Each DD method is then strictly compared against the \zi{} control generated from its own starting lines, 
so any difference isolates the effect of pixel optimisation.
\subsection{Adapting GM, DM and MTT to CTC}
\label{sec:dd}
Let $\ell(\theta;\mathcal{B})$ be the CTC loss of the CRNN with parameters $\theta$ on a batch $\mathcal{B}$, each
line's negative log-likelihood divided by its target length and averaged over the batch. GM and DM iterate through the synthetic
lines sequentially: at iteration $t$ they update line $j=t \bmod K$ (round-robin, as the original methods iterate over
classes) against a real batch $\mathcal{B}_j$. This real batch is drawn either from the corresponding pseudo-class $\mathcal{C}_j$ (conditional matching) or from the
whole pool (unconditional matching). Both modes ensure that each synthetic line receives the exact same number of updates.

\paragraph{GM-CTC (gradient matching).}
Following~\cite{zhao2021dc}, we compute $g^{r}=\nabla_\theta\ell(\theta;\mathcal{B}_j)$ and
$g^{s}=\nabla_\theta\ell(\theta;\{(s_j,\tilde y_j)\})$, and the synthetic pixels are updated to minimise the distance metric used in the original formulation:
\begin{equation}
D(g^r,g^s)=\sum_{l\,:\,\dim\ge2}\;\sum_{i}\Bigl(1-\frac{\langle g^r_{l,i},g^s_{l,i}\rangle}{\lVert g^r_{l,i}\rVert\,\lVert g^s_{l,i}\rVert+10^{-6}}\Bigr),
\label{eq:dc}
\end{equation}
where $g_{l,i}$ is row $i$ (one output unit) of the gradient of parameter tensor $l$, and one-dimensional tensors
(such as biases and normalisation parameters) are ignored. Importantly, differentiating $D$ w.r.t.\ $s_j$ requires the second derivative of the
CTC loss, which PyTorch's fused CTC kernel lacks. To circumvent this, the synthetic branch relies on a custom CTC implementation written with differentiable tensor
operations (forward recursion using \texttt{logsumexp}). As in the original method, the network parameters $\theta$ are iteratively trained on synthetic batches between pixel updates and are periodically re-initialised at random. 
Frequencies and hyperparameters are given in \cref{sec:protocol}.

\paragraph{DM-CTC (distribution matching).}
Following~\cite{zhao2023dm}, we utilise a randomly initialised and frozen CNN $\psi $, which is periodically re-drawn during optimisation. A text line is
embedded by averaging its feature columns over its width:
$h(x)=\frac{1}{T'}\sum_{\tau=1}^{T'}\psi(x)_{:,\tau}\in\mathbb{R}^{128}$, and the pixels minimise
\begin{equation}
\mathcal{L}_{\mathrm{DM}}=\Bigl\lVert \tfrac{1}{|\mathcal{B}_j|}\textstyle\sum_{x\in\mathcal{B}_j}h(x)-h(s_j)\Bigr\rVert_2^2 .
\label{eq:dm-ctc}
\end{equation}

\paragraph{MTT-CTC (trajectory matching).}
Following~\cite{cazenavette2022mtt}, expert recognizers trained on the full pool store their parameters
$\theta^{*}_{e,t}$ after every epoch $t$. At each iteration we sample an expert and a start epoch $t\le t_{\max}$, initialise a student network at $\theta^{*}_{e,t}$ and take $n=20$ gradient steps
$\hat\theta_{k+1}=\hat\theta_{k}-\alpha\nabla_\theta\ell(\hat\theta_k;\mathcal{S}_k)$ on random mini-batches
$\mathcal{S}_k$ of synthetic lines (matching the experts' batch size). This inner loop relies on the differentiable CTC and a learnable step $\alpha$. The synthetic pixels and $\alpha$ are then updated to minimise:
\begin{equation}
\mathcal{L}_{\mathrm{MTT}}=\frac{\lVert\hat\theta_{n}-\theta^{*}_{e,t+1}\rVert_2^2}{\lVert\theta^{*}_{e,t}-\theta^{*}_{e,t+1}\rVert_2^2},
\label{eq:mtt}
\end{equation}
\ie $n$ synthetic steps must reach where the expert arrived after one real epoch. To maintain numerical stability, updates resulting in non-finite losses or gradients are skipped. Full details regarding expert pre-training, specific values for $t_{\max}$ and $n$, bounds for $\alpha$, and gradient clipping are provided in \cref{sec:protocol}.

%% file: sections/setup.tex
\section{Experimental setup}
\label{sec:setup}

\begin{table*}[t]
\centering
\small
\caption{Datasets: lines per split as used (\cref{sec:data}; LAM train excludes one line too long for CTC to align), alphabet size $|\Sigma|$, and subset size $K$ per budget.}
\label{tab:datasets}
\setlength{\tabcolsep}{4.5pt}
\begin{tabular}{@{}llllrrrrr@{}}
\toprule
Dataset & Language & Period & Writers & Train & Val & Test & $|\Sigma|$ & $K$ (1/5/10\,\%) \\
\midrule
\input{tables/datasets_rows.tex} 
\bottomrule
\end{tabular}
\end{table*}

\subsection{Datasets and Preprocessing}
\label{sec:data}
We evaluated four line-level HTR datasets that differ in language, period, number of writers and size
(\cref{tab:datasets}): IAM~\cite{marti2002iam} (modern English, many writers; using the writer-independent line task, with
its two validation sets merged), George Washington (GW)~\cite{fischer2012lexicon} (18th-century English, two writers); READ\,2016~\cite{sanchez2016icfhr} (Early Modern German council
minutes, several hands; using the competition split) and LAM~\cite{cascianelli2022lam}
(Italian letters of a single writer).
\paragraph{Preprocessing.} All text-line images were converted to greyscale ($[0,255]$) and resized to a fixed height of $H=64$\,px while strictly preserving their original aspect ratio. The target transcriptions were defined over a dataset-specific character alphabet $\Sigma$.
\subsection{Training and evaluation protocol}
\label{sec:protocol}
\paragraph{Architecture.} As introduced in \cref{sec:crnn}, we used a small CRNN comprising four blocks of $5{\times}5$ convolutions (32 channels, no bias), batch normalisation, LeakyReLU (slope 0.2), and max-pooling ($2{\times}2$ three times, then $2{\times}1$). The resulting 128 features fed a one-layer bidirectional LSTM (64 units per direction) with a dropout of 0.5. To prevent an early all-blank solution, the output-layer bias for the CTC blank symbol was initialised to $-5$.
\paragraph{Base training.} Every experimental configuration, including the full-data reference, trained this CRNN from scratch using Adam (learning rate $3\cdot10^{-3}$) for at most 4{,}000 epochs. We validated every 20 epochs and applied early stopping after 30 validations without an improvement of 0.01 in validation \cer; the checkpoint with the best validation \cer{} was tested. On synthetic subsets, each optimiser step used all $K$ lines (processed in mini-batches of 32 with gradient accumulation), whereas the full pool used standard mini-batches of 32. Both real images and synthetic tensors were augmented with a random affine transform (rotation ${\pm}3^\circ$, translation ${\pm}5\,\%$, probability 0.5). We ran experiments with seeds 0, 42, and 1337 (adding 7 and 123 for GW to reduce variance on this tiny dataset). For \emph{Random} lines, the seed determined a different initial subset; for \emph{Text medoids}, the initial subset was fixed, so the seed only changed the training dynamics. \cer{} was calculated as the total edit distance divided by the total number of reference characters in the split (spaces included) using greedy CTC decoding. We report test \cer{} (model selection used validation only) and provide validation results in the supplement.
\paragraph{Distillation hyperparameters.} We detail here the specific parameters extracted from the original distillation methods (\cref{sec:init,sec:dd}):
\begin{itemize}
    \item \textbf{Initialisation:} Text pseudo-classes were clustered using $k$-means with 10 restarts.
    \item \textbf{GM-CTC:} The network parameters $\theta$ were trained for 5 Adam steps (learning rate $3\cdot10^{-3}$) on synthetic batches every 10 iterations, and were randomly re-initialised every 200 iterations.
    \item \textbf{DM-CTC:} The randomly initialised and frozen feature extractor $\psi$ was re-drawn every 20 iterations.
    \item \textbf{MTT-CTC:} Expert networks were pre-trained using Adam (learning rate $10^{-3}$), batches of 8 lines, and the standard augmentation described above. During distillation, the maximum expert start epoch was set to $t_{\max}=60$ for IAM, and $40$ for the others. The student took $n=20$ gradient steps per iteration using a learnable step size $\alpha$ (initialised to $0.01$ and clamped to $[10^{-4},0.1]$). Gradients were clipped to a maximum norm of 10.
\end{itemize}

\subsection{Distillation settings and cost}
\label{sec:ddsettings}
Unless stated otherwise, all distillation methods ran for 2{,}000 iterations and pixel learning rate 0.1. MTT used 15 expert trajectories per dataset (limited to 5 on LAM due to computational constraints). Regarding computational cost measured on a single GPU, GM and DM took 0.20--0.66\,h per experiment cell. Generating one MTT buffer takes 5.8--11.6\,h (median per cell), in addition to the 0.6--13.6\,h per dataset needed to pre-train the expert trajectories. Finally, training a single recognizer took 0.02--4.9\,h on a data subset, compared to 0.4--17.2\,h on the full training pool.

\subsection{Statistics}
\label{sec:stats}
We call each dataset--budget combination a \emph{cell} (4 datasets $\times$ 3 budgets $=12$ cells). In every cell,
each configuration is trained with $n=3$ seeds ($n=5$ on GW), and we report the mean $\pm$ 95\,\% confidence interval
(Student $t$; $\pm4.30$ standard errors for $n=3$).

\emph{What we compare.} Our main question is whether optimising the pixels helps. The \emph{primary} comparison is
thus each DD method against the \zi{} control built from its own starting lines, with both sides trained through
the same pipeline on the same GPU type and paired by seed. \emph{Secondary} comparisons set the methods against the
Random and Text-medoid baselines and compare starting lines with each other (paired by seed when both sides use the same lines and Welch's $t$ otherwise). Conditioning comparisons (conditional \vs unconditional matching) are paired by seed.

\emph{Combining cells.} For each comparison we take the mean difference in each of the 12 cells and test, with a
two-sided Wilcoxon signed-rank test, whether these differences consistently lean one way. This gives one $p$-value
per comparison, Holm-corrected within its family (5 primary, 11 secondary, 2 conditioning).

\emph{When we call it an effect.} A difference counts as an effect if its 95\,\% interval excludes 0 in a cell, or if
the Holm-corrected test across cells is significant at 0.05.



%% file: tables/datasets_rows.tex
IAM~\cite{marti2002iam} & English & modern & 657 & 6{,}161 & 1{,}840 & 1{,}861 & 79 & 62/308/616 \\
GW~\cite{fischer2012lexicon} & English & 18th c. & 2 & 357 & 68 & 68 & 70 & 4/18/36 \\
READ\,2016~\cite{sanchez2016icfhr} & German & 1470--1805 & several & 8{,}366 & 1{,}037 & 1{,}134 & 90 & 84/418/835 \\
LAM~\cite{cascianelli2022lam} & Italian & 1691--1750 & 1 & 19{,}829 & 2{,}470 & 3{,}523 & 89 & 198/991/1983 \\

%% file: sections/results.tex
\section{Results}
\label{sec:results}
We first compare the distilled sets against real lines of the same size (\cref{sec:rq1}), then investigate the source of their accuracy (\cref{sec:rq2}), whether stronger optimisation improves performance (\cref{sec:strength}) and whether the methods succesfully minimise their
own objectives (\cref{sec:objective}). Finally, \cref{sec:pitfalls} discusses several practical pitfalls.
\begin{table*}[t]
\centering
\footnotesize
\caption{Test \cer{} (\%, mean $\pm$ 95\,\% CI; $n=3$, GW $n=5$; one decimal). Each block starts from the same $K$ real
lines: the first column trains on them as images (the selection baseline), \zi{} trains on the same lines as tensors
without optimisation (the control of the DD columns), and the DD columns optimise their pixels. Random: a different
subset per seed. Nothing is bold: tested differences are in \cref{tab:comparisons}. A single random subset fixed
across seeds is reported in the supplement.}
\label{tab:main}
\setlength{\tabcolsep}{2.8pt}
\begin{tabular}{@{}llccccccccc@{}}
\toprule
& & \multicolumn{5}{c}{Random lines} & \multicolumn{4}{c}{Text medoids} \\
\cmidrule(lr){3-7}\cmidrule(l){8-11}
Dataset & Budget ($K$) & Random & \zi & DM & GM & MTT & Medoids & \zi & DM$_\mathrm{med}$ & GM$_\mathrm{med}$ \\
\midrule
\input{tables/main_test.tex} 
\bottomrule
\end{tabular}

\vspace{2pt}
\raggedright\footnotesize Full training pool (reference, see \cref{sec:protocol}): \input{tables/full_test}
\end{table*}

\begin{figure}[t]
\centering
\includegraphics[width=\linewidth]{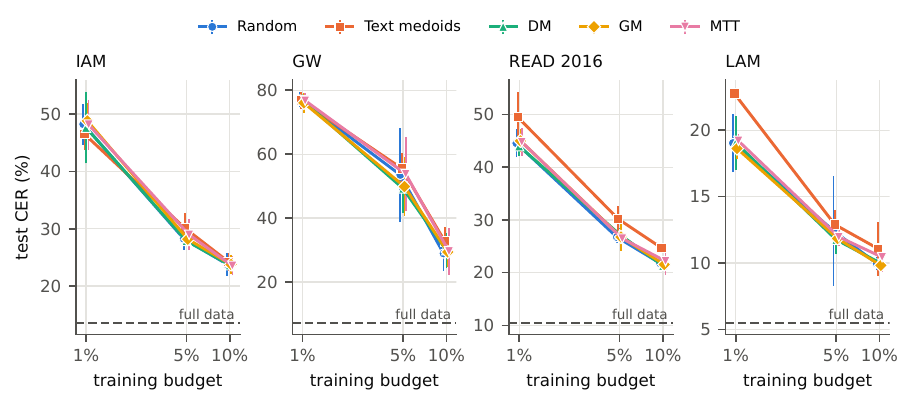}
\caption{Test \cer{} against budget (mean, 95\,\% CI; points offset for legibility). Random: one subset per seed; DM,
GM and MTT start from the same random lines. Dashed: full training pool.}
\label{fig:budget}
\end{figure}

\subsection{Distillation does not beat real lines of the same size}
\label{sec:rq1}
As shown in \cref{tab:main} and \cref{fig:budget}, the methods yield overlapping performance at every budget and their relative ranking varies across experimental cells. When initialised with random lines, DM, GM and MTT perform comparably to the \emph{Random} baseline in all cells (\cref{tab:comparisons}; Wilcoxon
\tc{$p=0.176$}, \tc{$0.301$}, \tc{$0.339$}). When initialised with \emph{Text medoids}, DM$_\mathrm{med}$ and GM$_\mathrm{med}$
never outperform the \emph{Random} baseline and are strictly worse in \tc{4} of 12 cells (READ\,2016, LAM 1\,\%; Holm \tc{$0.161$},
\tc{$0.0742$}). 
Finally, even at a 10\,\% budget, every method falls significantly short of the model trained on the full pool (e.g., on IAM: \tc{23.5}--\tc{24.6} \vs \tc{13.6}).

\begin{table}[t]
\centering
\scriptsize
\caption{Comparisons over the 12 cells (test): Wilcoxon $p$ on per-cell mean differences, Holm-corrected per family,
and cells whose 95\,\% CI excludes 0 with A better or worse. DM, GM, MTT: from the random lines; DM$_\mathrm{med}$,
GM$_\mathrm{med}$: from the text medoids; 0-iter: the starting lines of the method without optimisation; Random: one subset per seed. 
}
\label{tab:comparisons}
\setlength{\tabcolsep}{3pt}
\begin{tabular}{@{}lcccc@{}}
\toprule
A $-$ B & $p$ & Holm $p$ & A better & A worse \\
\midrule
\input{tables/comparisons.tex} 
\bottomrule
\end{tabular}
\end{table}

\subsection{A distilled set performs like its starting lines}
\label{sec:rq2}

\begin{figure}[t]
\centering
\includegraphics[width=\linewidth]{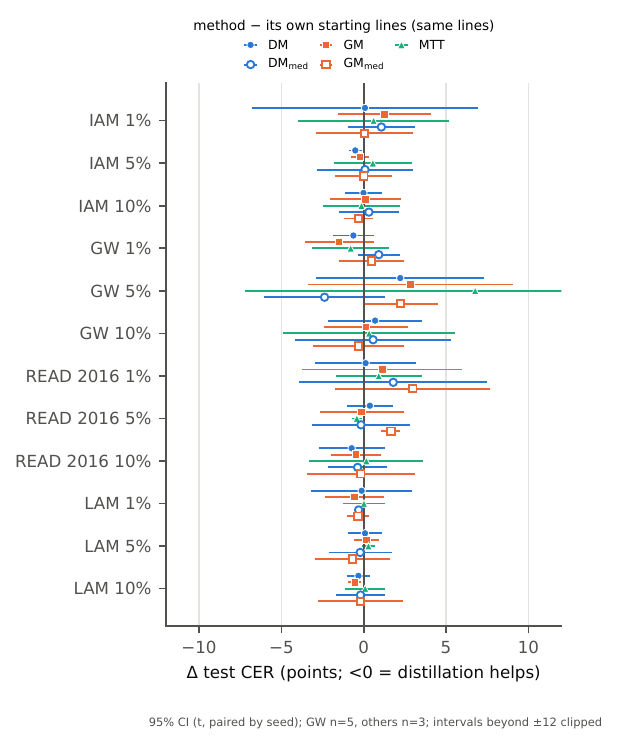}
\caption{Difference in test \cer{} between each distilled set and its starting lines trained without optimisation
(paired by seed; mean and 95\,\% CI). Filled: DM, GM and MTT from random lines; open: DM$_\mathrm{med}$ and
GM$_\mathrm{med}$ from the medoids. Negative: distillation helps.}
\label{fig:delta}
\end{figure}

\paragraph{No method improves upon its starting lines.}
We compare each DD method against the \zi{} control built from its exact starting lines (\cref{tab:comparisons},
\cref{fig:delta}). Since both models undergo the identical tensor training pipeline on the same GPU type, any difference stems from pixel optimisation. Across the 12 experimental cells, none of the five methods differs from its starting lines
(Wilcoxon signed-rank test): when initialised from random lines, DM \tc{$p=0.97$}, GM \tc{$p=1$} and MTT \tc{$p=0.151$}; from Text medoids,
DM$_\mathrm{med}$ \tc{$p=0.677$} and GM$_\mathrm{med}$ \tc{$p=0.733$}. After Holm correction across these five primary comparisons, all differences remain non-significant (\tc{$p\ge0.757$}).

An analysis of individual cells corroborates this finding. Out of the 60 method--cell intervals, only 5 exclude 0, which aligns closely with the three false positives expected by chance at a 95\,\% confidence level. Four of these instances represent marginal improvements of at most \tc{0.53} points, whereas the fifth constitutes a degradation (GM$_\mathrm{med}$ at READ\,2016 5\,\%, \tc{$+1.65\pm0.58$}). Given that $n=3$, most confidence intervals remain relatively wide (frequently spanning ${\pm}2$ points); thus, while these findings definitively rule out large performance gains, we cannot discard the possibility of negligible, small-scale effects.

\paragraph{Minimal pixel displacement under default settings.}
The optimisation process induces negligible modifications to the synthetic text lines. Regardless of the starting lines, the relative change $\lVert S-S_0\rVert_2/\lVert S_0\rVert_2$ remains between 0.05\,\% and 0.39\,\% for DM and GM across most datasets (0.4--3.7\,\% on GW), and ranges from 0.21\,\% to 2.83\,\% for MTT. This marginal displacement aligns exactly with the bounds imposed by the step size: each gradient update shifts a pixel by ${\lesssim}\,0.1$ grey levels, and each line is subjected to only 1--32 updates during the entire distillation process (56--500 on the exceptionally small GW dataset).
\paragraph{Performance is driven by the starting lines.}
\label{sec:init-matters}
The only meaningful differences we observe stem from the choice of starting lines, where \emph{Text medoids} prove to be the weaker option. They never outperform the \emph{Random} baseline and are strictly worse in \tc{4} cells: READ\,2016 across all budgets and LAM 1\,\% (Wilcoxon
\tc{$p=0.00684$}, Holm \tc{$0.0684$}).
Pixel distillation preserves this performance gap. DM$_\mathrm{med}$ is strictly worse than DM in \tc{5} cells (Wilcoxon \tc{$p=0.000977$}, Holm
\tc{$p=0.0107$}), representing the only Holm-significant comparison in the entire study. GM$_\mathrm{med}$ exhibits the same trend, underperforming GM in \tc{5} cells (Holm \tc{$0.0742$}).
Crucially, this gap is not attributable to conditional matching. While the medoid variants match each line only against real lines from its own
cluster, matching them instead against batches drawn from the whole pool (while maintaining the exact same number of updates per line),
yields no measurable difference (\cref{tab:comparisons}, conditioning family): DM yields \tc{$p=0.91$} (with one cell favouring
conditioning), and GM
yields \tc{$p=0.569$} (with none; Holm \tc{$1$} for both). Although preliminary runs initially suggested a conditioning effect, this was an artefact of the unconditional loop updating up to 32 lines per iteration instead of one; the effect vanished once the number of updates per line was equalised across both loops.
This finding carries an important methodological implication: evaluating a DD method exclusively against a non-random selection can artificially inflate its perceived effectiveness. For instance, MTT outpeforms \emph{Text medoids} in \tc{4} cells (Holm \tc{$p=0.0684$}), yet this advantage stems entirely from its random initialisation, since when compared directly against the Random baseline, MTT remains at parity in every cell.

\subsection{Does stronger optimisation improve performance?}
\label{sec:strength}
\begin{figure}
\centering
\includegraphics[width=\linewidth]{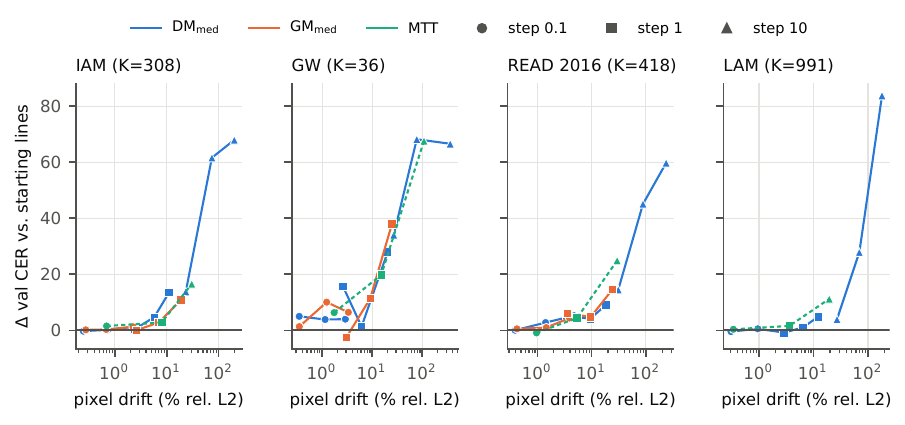}
\caption{Optimisation strength (one seed, validation): change in \cer{} with respect to the starting lines against
pixel drift (relative $\ell_2$, log scale) for DM$_\mathrm{med}$ and GM$_\mathrm{med}$ (batch mode;
$U\approx30,300,3000$ updates per line along each line) and MTT, by pixel step. Moving the pixels away from the
starting lines increases the error; step 10 collapses the recognizer.}
\label{fig:drift}
\end{figure}

\begin{figure}[t]
\centering
\includegraphics[width=\linewidth]{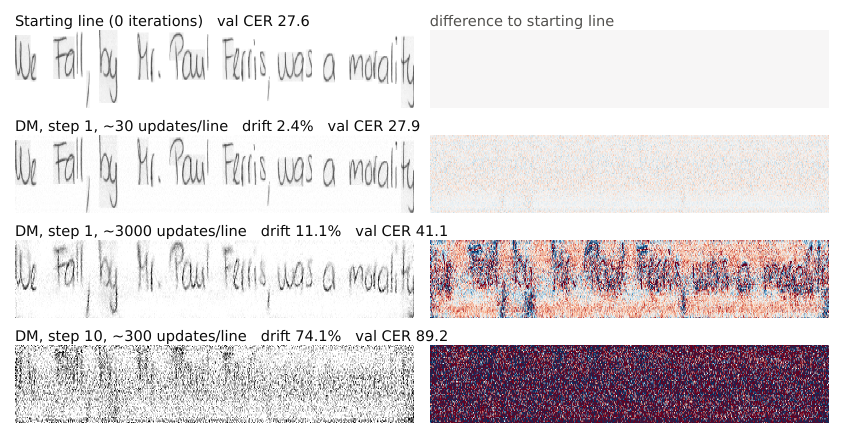}
\caption{An IAM line ($K=308$) at its starting point and after DM$_\mathrm{med}$ in batch mode with increasing optimisation strength
(left), and its difference to the starting line (right). Drift and validation \cer{} (one seed) refer to the whole
buffer.}
\label{fig:examples}
\end{figure}

Since the default settings induce minimal pixel modifications, the null result could indicate that the optimisation was too weak.
On one budget per dataset (5\,\%; 10\,\% for GW, whose 5\,\% cell exhibited high variance), we ran DM$_\mathrm{med}$ and
GM$_\mathrm{med}$ in batch mode (32 synthetic lines per iteration, unconditional), varied the pixel step size
$\eta\in\{0.1,1,10\}$ and took snapshots at $U\approx30$, 300 and 3{,}000 updates per line. Phase A selected the best
$(\eta,U)$ on validation with one seed, and Phase B retrained it with three seeds. For MTT, we varied only the pixel learning
rate (one seed). \Cref{fig:drift} shows the change in validation \cer{} against the resulting drift and \cref{fig:examples}
displays some example lines.

\paragraph{DM and GM.}
In Phase A, no configuration clearly outperforms the starting lines, and the error increases with both $\eta$ and $U$.
On IAM ($K=308$; \zi{} $27.39\pm0.84$) the best DM configuration is the one with minimal pixel drift ($\eta=0.1$, $U\approx30$:
$27.36$); with $\eta=1$ the \cer{} rises from $27.95$ to $32.32$ and $41.14$ as $U$ increases, and with $\eta=10$ the recognizer
collapses ($41.47$, $89.24$, $95.60$). The remaining datasets and GM follow the same pattern; on GW, no configuration reaches
the \zi{} performance level (best: $30.20$ against $27.26\pm3.85$). When retrained with three seeds (Phase B), the best configuration for
each method and dataset remains statistically indistinguishable from its \zi{} control in all 7 cases, both on validation ($\Delta$ from
$-0.52\pm1.16$ to $+4.22\pm12.94$) and on test (\tc{$-0.56\pm1.37$} to \tc{$+4.02\pm12.19$}).

\paragraph{MTT.}
Increasing MTT's pixel learning rate from 0.1 to 1 and 10 monotonically degrades validation \cer{} across all four datasets
(e.g., on IAM: $28.76\to30.00\to43.78$, compared to $27.19$ for the starting lines).

\subsection{Do the methods lower their own objective?}
\label{sec:objective}
Parity with the starting lines could indicate either that minimising the objective does not improve performance, or that it was never effectively optimised.
For each run, we evaluated its own loss on its starting and final lines under the same networks and real
batches, and report the final-to-initial ratio (95\,\% CI; per-run values in the supplement).
\paragraph{DM lowers its objective, without benefit.}
Across all 24 runs, the final lines yield a lower DM loss: by $0.1$--$2.5$\,\% outside GW (ratio $0.975$--$0.999$) and by
8--30\,\% on GW 1 and 5\,\% ($0.70$--$0.92$), where each line receives 111--500 updates. Minimising it further does not improve performance:
in the IAM sweep at $\eta=0.1$, the objective decreases by 8.5\,\% ($0.888$--$0.944$) after $U\approx3000$ updates
while validation \cer{} rises to $28.89$ (starting lines: $27.39$), and with larger step sizes the objective increases
(1.08--8.7).
\paragraph{GM barely moves its objective.}
Across the 24 runs and four network states (freshly initialised, and after 20, 50 and 100 co-adaptation steps), the ratio remains between
$0.969$ and $1.002$ (median $0.9996$): at most 0.5\,\% lower outside GW and 3\,\% on GW, never reaching 5\,\%. Applying  thousands of updates per
line in the IAM sweep do not change this ($0.990$--$1.008$ at $\eta=0.1$), whereas larger step sizes increase it (up to 1.44 at
$\eta=1$).
\paragraph{The MTT target is unreachable.}
The MTT loss (\cref{eq:mtt}) ends at 0.979--0.999 (last 500 iterations across 36 runs), virtually unchanged from its initial value. Running the same
20-step inner loop on real lines of the full pool, using the best step size from a grid ($\alpha\in[0.003,3]$; larger step sizes
overshoot), yields the same values (median 0.997 on IAM, 0.981 on GW, 0.998 on READ\,2016, 0.999 on LAM): neither real nor synthetic lines bring the student close to the expert's next epoch. The expert's one-epoch displacement is nearly
orthogonal to the mean training gradient (cosine similarity 0.02--0.20), consistent with reports that Adam or momentum experts
degrade MTT in image classification~\cite{shen2023ast}. On IAM, using SGD experts (as in the original method) makes the target
reachable only early in training or when they converge poorly (validation \cer{} 21--40 after 20 epochs \vs 17 with
Adam; supplement). For GM and MTT, the null result of \cref{sec:rq2} thus reflects an objective that the optimisation
barely alters or cannot reach, not evidence against the matching principle itself.
\subsection{Practical pitfalls}
\label{sec:pitfalls}
\paragraph{A single random subset is a fragile reference at low budgets.}
At 1\,\%, using a different random subset per seed reaches $50.68\pm3.93$ validation \cer{} on READ\,2016 against
$56.65\pm4.64$ for a single subset fixed across seeds (test \tc{$44.6\pm2.6$} \vs \tc{$49.8\pm4.1$}), and
$47.00\pm3.71$ against $49.67\pm5.18$ on IAM; on GW and LAM both strategies align. 
The READ\,2016 gap stems from sampling variance, which is as large as the differences between methods. 
Against that single subset MTT appeared superior (\tc{$44.8\pm2.6$}), yet its own starting lines already reached \tc{$43.9\pm2.7$}. We therefore use one
random subset per seed as the \emph{Random} baseline, and report the fixed subset in the supplement (\tc{$p=0.0771$} between the
two).
\paragraph{Second-order CTC gradients are fragile.}
In batch mode, GM skipped 0.9--2.6\,\% of pixel updates (non-finite values) \vs none with one line per iteration (supplement).

\paragraph{Infinite-loss lines break MTT silently.}
Whenever any of the starting lines produce non-finite loss or gradients, MTT fails an iteration. This was observed in the seed-0 runs of IAM 5 and 10\,\%, which skipped 41.5\,\% and 23.6\,\% of iterations against the $1-(1-8/K)^{20}=40.9\,\%$ and 23.0\,\% predicted. Skips on LAM, where all losses and gradients are finite, remain unexplained (supplement).

%% file: tables/main_test.tex
IAM & 1\,\% (62) & \tc{48.2$\pm$3.6} & \tc{47.5$\pm$1.2} & \tc{47.6$\pm$6.2} & \tc{48.8$\pm$3.2} & \tc{48.2$\pm$4.4} & \tc{46.5$\pm$2.7} & \tc{46.3$\pm$1.0} & \tc{47.4$\pm$1.1} & \tc{46.4$\pm$3.2} \\
 & 5\,\% (308) & \tc{28.3$\pm$1.9} & \tc{28.4$\pm$0.5} & \tc{27.9$\pm$0.2} & \tc{28.2$\pm$0.5} & \tc{29.0$\pm$2.7} & \tc{30.1$\pm$2.7} & \tc{29.4$\pm$1.5} & \tc{29.5$\pm$1.5} & \tc{29.4$\pm$0.5} \\
 & 10\,\% (616) & \tc{23.8$\pm$2.0} & \tc{23.7$\pm$0.8} & \tc{23.7$\pm$0.8} & \tc{23.8$\pm$1.7} & \tc{23.5$\pm$1.7} & \tc{24.2$\pm$1.6} & \tc{24.3$\pm$0.7} & \tc{24.6$\pm$1.2} & \tc{24.0$\pm$0.6} \\
\midrule
GW & 1\,\% (4) & \tc{77.1$\pm$2.4} & \tc{77.5$\pm$1.3} & \tc{76.9$\pm$2.0} & \tc{76.0$\pm$3.0} & \tc{76.7$\pm$2.3} & \tc{76.8$\pm$2.7} & \tc{77.1$\pm$1.5} & \tc{78.0$\pm$2.5} & \tc{77.5$\pm$2.4} \\
 & 5\,\% (18) & \tc{53.5$\pm$14.8} & \tc{47.0$\pm$5.5} & \tc{49.3$\pm$7.6} & \tc{49.9$\pm$9.1} & \tc{53.8$\pm$11.6} & \tc{55.6$\pm$4.7} & \tc{54.6$\pm$4.4} & \tc{52.2$\pm$2.6} & \tc{56.8$\pm$4.2} \\
 & 10\,\% (36) & \tc{28.9$\pm$5.4} & \tc{29.3$\pm$2.9} & \tc{30.0$\pm$5.5} & \tc{29.4$\pm$4.8} & \tc{29.6$\pm$7.4} & \tc{32.7$\pm$4.5} & \tc{32.0$\pm$3.2} & \tc{32.6$\pm$5.7} & \tc{31.7$\pm$4.2} \\
\midrule
READ\,2016 & 1\,\% (84) & \tc{44.6$\pm$2.6} & \tc{43.9$\pm$2.7} & \tc{44.0$\pm$1.9} & \tc{45.0$\pm$2.2} & \tc{44.8$\pm$2.6} & \tc{49.5$\pm$4.8} & \tc{49.0$\pm$5.3} & \tc{50.8$\pm$4.7} & \tc{52.0$\pm$5.2} \\
 & 5\,\% (418) & \tc{26.7$\pm$1.1} & \tc{26.9$\pm$0.9} & \tc{27.3$\pm$1.1} & \tc{26.8$\pm$2.6} & \tc{26.5$\pm$0.7} & \tc{30.2$\pm$2.5} & \tc{29.9$\pm$1.5} & \tc{29.7$\pm$1.7} & \tc{31.5$\pm$2.1} \\
 & 10\,\% (835) & \tc{21.8$\pm$0.8} & \tc{22.0$\pm$1.5} & \tc{21.3$\pm$0.6} & \tc{21.6$\pm$0.5} & \tc{22.2$\pm$2.6} & \tc{24.7$\pm$0.4} & \tc{24.2$\pm$2.3} & \tc{23.9$\pm$0.5} & \tc{24.1$\pm$2.2} \\
\midrule
LAM & 1\,\% (198) & \tc{19.0$\pm$2.2} & \tc{19.2$\pm$1.1} & \tc{19.1$\pm$2.0} & \tc{18.6$\pm$0.8} & \tc{19.2$\pm$0.5} & \tc{22.8$\pm$0.1} & \tc{23.5$\pm$1.0} & \tc{23.2$\pm$0.8} & \tc{23.2$\pm$0.6} \\
 & 5\,\% (991) & \tc{12.4$\pm$4.1} & \tc{11.7$\pm$0.5} & \tc{11.8$\pm$1.1} & \tc{11.9$\pm$0.6} & \tc{12.0$\pm$0.7} & \tc{12.9$\pm$1.1} & \tc{13.1$\pm$2.0} & \tc{12.9$\pm$1.7} & \tc{12.4$\pm$0.9} \\
 & 10\,\% (1983) & \tc{10.0$\pm$0.5} & \tc{10.4$\pm$0.6} & \tc{10.0$\pm$0.3} & \tc{9.8$\pm$0.3} & \tc{10.5$\pm$1.1} & \tc{11.1$\pm$2.0} & \tc{11.0$\pm$1.2} & \tc{10.8$\pm$2.3} & \tc{10.8$\pm$1.4} \\

%% file: tables/full_test.tex
IAM \tc{13.6$\pm$0.5}; GW \tc{7.2$\pm$1.1}; READ\,2016 \tc{10.5$\pm$0.8}; LAM \tc{5.5$\pm$0.1}

%% file: tables/comparisons.tex
\multicolumn{5}{@{}l}{\emph{Primary (paired, own init)}} \\
\quad DM $-$ 0-iter & \tc{0.97} & \tc{1} & \tc{1} & \tc{0} \\
\quad GM $-$ 0-iter & \tc{1} & \tc{1} & \tc{1} & \tc{0} \\
\quad MTT $-$ 0-iter & \tc{0.151} & \tc{0.757} & \tc{1} & \tc{0} \\
\quad DM$_\mathrm{med}$ $-$ 0-iter & \tc{0.677} & \tc{1} & \tc{1} & \tc{0} \\
\quad GM$_\mathrm{med}$ $-$ 0-iter & \tc{0.733} & \tc{1} & \tc{0} & \tc{1} \\
\multicolumn{5}{@{}l}{\emph{Secondary (paired if same lines, else Welch)}} \\
\quad DM $-$ Random & \tc{0.176} & \tc{0.881} & \tc{0} & \tc{0} \\
\quad GM $-$ Random & \tc{0.301} & \tc{1} & \tc{0} & \tc{0} \\
\quad MTT $-$ Random & \tc{0.339} & \tc{1} & \tc{0} & \tc{0} \\
\quad Medoids $-$ Random & \tc{0.00684} & \tc{0.0684} & \tc{0} & \tc{4} \\
\quad DM$_\mathrm{med}$ $-$ Random & \tc{0.0269} & \tc{0.161} & \tc{0} & \tc{4} \\
\quad GM$_\mathrm{med}$ $-$ Random & \tc{0.00977} & \tc{0.0742} & \tc{0} & \tc{4} \\
\quad DM$_\mathrm{med}$ $-$ DM & \tc{0.000977} & \tc{0.0107} & \tc{0} & \tc{5} \\
\quad GM$_\mathrm{med}$ $-$ GM & \tc{0.00928} & \tc{0.0742} & \tc{0} & \tc{5} \\
\quad DM$_\mathrm{med}$ $-$ Medoids & \tc{0.97} & \tc{1} & \tc{1} & \tc{0} \\
\quad GM$_\mathrm{med}$ $-$ Medoids & \tc{0.677} & \tc{1} & \tc{0} & \tc{0} \\
\quad MTT $-$ Medoids & \tc{0.00684} & \tc{0.0684} & \tc{4} & \tc{0} \\
\multicolumn{5}{@{}l}{\emph{Conditioning (paired)}} \\
\quad DM$_\mathrm{med}$: cond.\ $-$ uncond. & \tc{0.91} & \tc{1} & \tc{1} & \tc{0} \\
\quad GM$_\mathrm{med}$: cond.\ $-$ uncond. & \tc{0.569} & \tc{1} & \tc{0} & \tc{0} \\

%% file: sections/discussion.tex
\section{Discussion}
\label{sec:discussion}

\paragraph{Why does distillation not help here?}
For GM and MTT, the answer is partly mechanical (\cref{sec:objective}): their objectives barely change, even with
thousands of updates per line (GM) or SGD experts (MTT). DM does minimise its objective, yet without benefit. We see three untested explanations:
(i) \emph{Fixed sequence labels.} A synthetic image of a class can drift towards a class prototype; a synthetic line
must remain readable as one specific character sequence, which leaves little room to add information it did not
already contain.
(ii) \emph{An uninformative embedding.} DM averages features over the width of the line and ignores character order.
(iii) \emph{Budget regime and model size.} Even 1\,\% of the lines contains many instances of most characters, and DD
loses its advantage over random selection as samples per class grow~\cite{cui2022dcbench}. Finally, a small recognizer may
also extract less from optimised pixels than a larger one.

\paragraph{What might work.}
Learning transcriptions or alignments with the pixels~\cite{bohdal2020flexible,cui2023tesla}, optimising a handwriting
generator's latent space~\cite{cazenavette2023glad,kang2020ganwriting}, and, for MTT, training experts whose trajectories are reachable, \ie whose next epoch a student on real lines can approach (\cref{sec:objective}).

\begin{figure}[t]
\centering
\fbox{\begin{minipage}{0.95\linewidth}\small
\textbf{Guidelines for evaluating data compression in HTR}
\begin{enumerate}[nosep,leftmargin=*]
\item Compare every distilled set with its \emph{own starting lines} trained without optimisation (same pipeline, recipe and GPU type).
\item Check that the method lowers its own objective (initial \vs final set, same networks and batches) and, for
  trajectory matching, that real data can reach the target.
\item Report Random over \emph{several subsets}; a selection baseline alone can make a method look better than it is.
\item Report pixel step relative to pixel range, updates per line and drift; match updates per line in ablations; report skipped non-finite updates.
\end{enumerate}
\end{minipage}}
\end{figure}

\paragraph{Limitations.}

Our study is restricted to a single small CRNN at the line level with fixed transcriptions. Most experiments used $n=3$ seeds ($n=5$ on GW), so per-cell intervals remain wide. Moreover, strength sweeps used a single seed, validation tuning was applied to DD only, and different GPU types shifted \cer{} by ${\sim}0.5$ (though never mixed within a test). Additionally, MTT on LAM used 5 experts; GM's objective was measured only on fresh or briefly co-adapted networks, and SGD experts only on IAM. 

%% file: sections/conclusion.tex
\section{Conclusion}
\label{sec:conclusion}
Adapted to CTC-based HTR, gradient, distribution and trajectory matching do not measurably improve upon the real lines they start from. A distilled set is only as good as its starting lines, and random lines serve as a stronger initialisation than text medoids.
We hope our controls help future data compression methods for HTR outperform their own starting point.

%% file: sections/ack.tex
\paragraph{Use of LLMs.}
LLMs helped implement and launch some experiments, cross-check reported numbers, and edit the text. The authors designed the study, checked all code, results and text, and take full responsibility for the paper.